%% file: MonoViT.tex
\documentclass[10pt,twocolumn,letterpaper]{article}

\usepackage{3dv}
\usepackage{times}
\usepackage{epsfig}
\usepackage{graphicx}
\usepackage{amsmath}
\usepackage{amssymb}
\usepackage{multirow}
\usepackage{threeparttable}
\usepackage{marvosym}

\usepackage[pagebackref=true,breaklinks=true,colorlinks,bookmarks=false]{hyperref}
\newcommand\blfootnote[1]{%
  \begingroup
  \renewcommand\thefootnote{}\footnote{#1}%
  \addtocounter{footnote}{-1}%
  \endgroup
}

\usepackage{pifont}

\threedvfinalcopy 

\def\threedvPaperID{69} 

\ifthreedvfinal\fi
\begin{document}

\title{MonoViT: Self-Supervised Monocular Depth Estimation with \\ a Vision Transformer}

\author{Chaoqiang Zhao$^{1,2,*}$ \hspace{0.5cm} Youmin Zhang$^{2,*}$ \hspace{0.5cm} Matteo Poggi$^{2}$ \hspace{0.5cm} Fabio Tosi$^{2}$ \hspace{0.5cm} Xianda Guo$^{3}$\\
Zheng Zhu$^{3}$ \hspace{0.5cm} Guan Huang$^{3}$ \hspace{0.5cm} Yang Tang$^1{} ^\ddag$ \hspace{0.5cm} Stefano Mattoccia$^{2}$\\
$^{1}$ East China University of Science and Technology \hspace{0.3cm}  $^{2}$ University of Bologna \hspace{0.3cm} $^{3}$ PhiGent Robotics \\
}


\twocolumn[{
\maketitle
\vspace{-1cm} 
\begin{center}
    \renewcommand{\tabcolsep}{1pt}
    \begin{tabular}{c c c}
    \includegraphics[height=.19\textwidth]{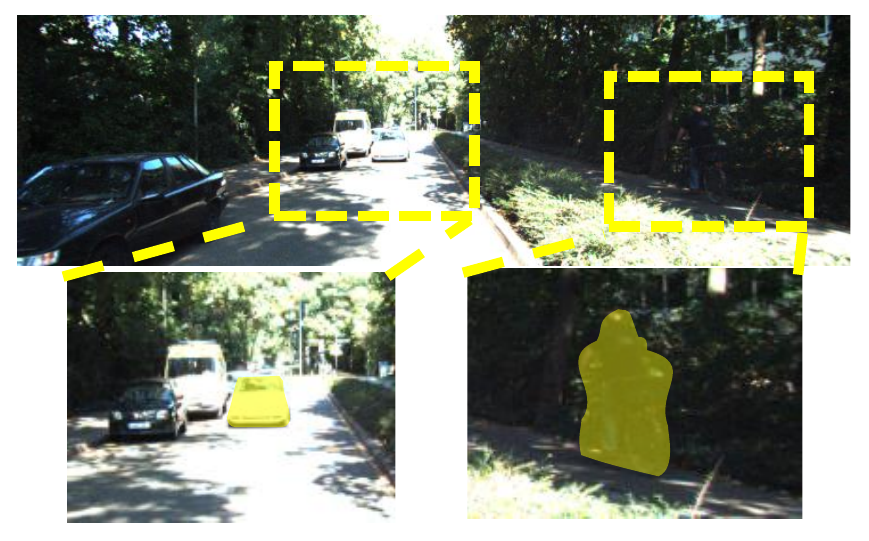} &
    \includegraphics[height=.189\textwidth]{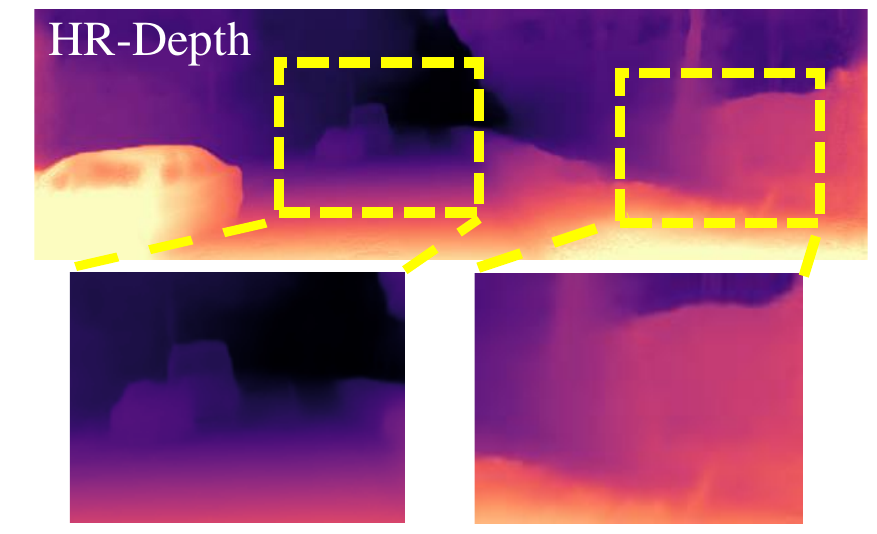} &
    \includegraphics[height=.188\textwidth]{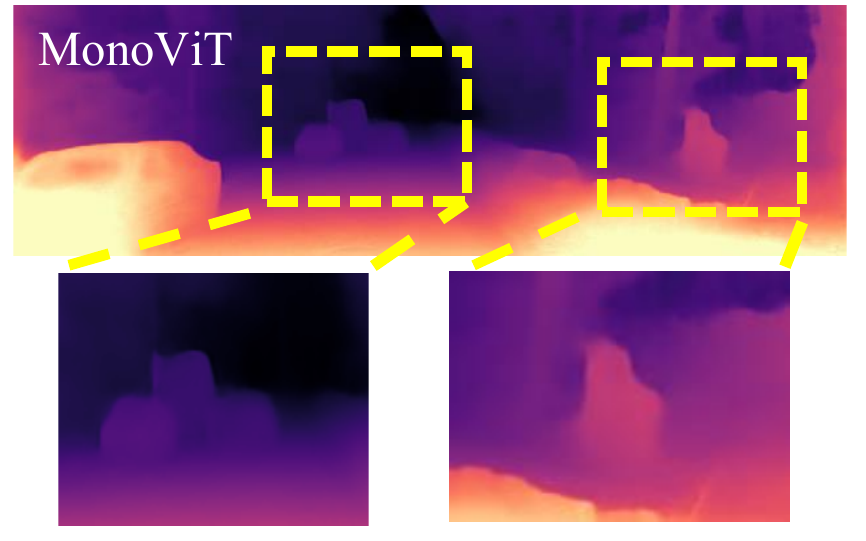} \\
    \end{tabular}
    \label{fig:ds}
\end{center}
\vspace{-0.2cm} 
\hypertarget{fig:ds}{Figure 1. }{\textbf{Effects of global reasoning on self-supervised monocular depth estimation.} The limited receptive field of existing solutions (\eg HR-Depth \cite{lyu2020hr}, in the middle) often yields inaccurate depth estimation, losing fine-grained details (like the car and cyclist over imposed in yellow). On the contrary, our MonoViT architecture (right) achieves superior results.}
\vspace{0.25cm}
}]


\begin{abstract} 
   Self-supervised monocular depth estimation is an attractive solution that does not require hard-to-source depth labels for training. Convolutional neural networks (CNNs) have recently achieved great success in this task. However, their limited receptive field constrains existing network architectures to reason only locally, dampening the effectiveness of the self-supervised paradigm. In the light of the recent successes achieved by Vision Transformers (ViTs), we propose MonoViT, a brand-new framework combining the global reasoning enabled by ViT models with the flexibility of self-supervised monocular depth estimation.
   By combining plain convolutions with Transformer blocks, our model can reason locally and globally, yielding depth prediction at a higher level of detail and accuracy, allowing MonoViT to achieve state-of-the-art performance on the established KITTI dataset. Moreover, MonoViT proves its superior generalization capacities on other datasets such as Make3D and DrivingStereo. Source code available at \url{https://github.com/zxcqlf/MonoViT}
\end{abstract}
\vspace{-0.28cm}
\section{Introduction}
\blfootnote{$^*$ Joint first authorship; $^\ddag$ Corresponding author.}Depth perception is at the foundation of several high-level computer vision applications such as autonomous driving, robotics, and augmented reality \cite{tang2022perception}. However, despite the steady progress of active depth-sensing technologies brought by devices such as LiDARs, Time-of-Flight (ToF) cameras and more, the possibility of estimating depth from standard images is generally preferable, mainly because of three (among many) advantages: higher image resolution, lower hardware costs and potentially unconstrained working range.
Although using two or more images~\cite{poggi2021synergies} is often the preferred choice, estimating depth from a single image allows for the deployment of depth-sensing solutions on any monocular configuration, still representing the most diffused setting in most practical cases nowadays.

Deep learning paradigms favoured the blooming of this latter approach~\cite{Eigen2014,laina2016,Fu2018,bhat2021adabins}, at the cost of requiring extensive collections of images annotated with depth labels in order to carry out the training effectively. 
However, considering the high cost of collecting dense depth labels for this purpose, self-supervised monocular depth estimation~\cite{Godard17,zhou_sfmlearner} has emerged in the literature enabling significant progress in recent years~\cite{zhao2020monocular}.
These approaches replace supervised losses on depth labels with supervisory signals derived from image reprojection across different views, by exploiting the geometric relationship between frames, \textit{i.e.} the scene depth itself and camera pose. 
Since these networks aim at learning depth, two prominent cases exist to deal with the relative pose across frames. They consist of either knowing it as a prior -- for instance, by collecting stereo images and training on them~\cite{Godard17} -- or estimating it during training, allowing in this second case to train on unconstrained monocular videos~\cite{zhou_sfmlearner}. The latter configuration turns out to be the preferred choice for practical deployment since it simply requires a single moving camera for gathering training data. For this reason, we stick to monocular videos for training purposes.
However, view reconstruction based losses suffer from occlusions, dynamic objects and photometric changes, which severely limit the performance of the network~\cite{zhao2020monocular}. Therefore, novel constraints~\cite{monodepth2} and additional cues~\cite{klingner2020self,yin2018geonet} (like semantic segmentation, optical flow and surface normals) are often used to reduce the shortcomings mentioned above.

Improving the network backbone of depth networks is another well-known effective way to gain accuracy. Recent research has shown that the encoder is crucial for achieving this~\cite{zhou_sfmlearner,monodepth2}. Different kinds of backbone, such as VGGNet, ResNet, HRNet and PackNet, made their way into the self-supervised monocular depth estimation task~\cite{zhou_sfmlearner,monodepth2,zhoudiffnet,guizilini2020}. Moreover, to improve the feature extraction and processing ability, new frameworks like HR-Depth~\cite{lyu2020hr} and CADepth~\cite{yan2021channel} also introduced attention modules. 
However, we argue that a shared shortcoming of existing self-supervised models falls in the reduced receptive field of Convolutional Neural Networks (CNNs). This fact represents an implicit bottleneck for current dense estimation methods, dampening accuracy, and the capacity to generalize to different domains. Specifically, the local nature of convolutions leads CNNs in their first layers -- \textit{i.e.}, those in charge of modeling fine-grained details -- to extract features missing long-range relationships across the same image. Going deeper with convolutions makes the receptive field wider, yet it does not reach the whole image. 
Fig.~\hyperref[fig:ds]{1} highlights the effect of this shortcoming. CNNs based frameworks sometimes fail to estimate foreground-background structures due to the lack of global perceiving and long-range relationship among modelled pixels. 
Vision Transformers (ViTs)~\cite{dosovitskiy2020image,xie2021segformer,dai2021up} recently showed outstanding results on tasks such as object detection~\cite{dai2021up} and semantic segmentation~\cite{xie2021segformer}, thanks to their capacity to model long-range relationships between pixels and thus a global receptive field. The popularity of ViTs has also reached supervised depth estimation as well~\cite{ranftl2021vision,li2022depthformer}, yet being not adopted for self-supervised monocular depth estimation. 

This paper takes this missing step and explores ViTs for self-supervised monocular depth estimation by proposing the MonoViT architecture. It combines both convolutional layers and state-of-the-art (SoTA) Transformer blocks~\cite{lee2021mpvit} within its backbone to model both the local information (objects) and global information (relationship among foreground and background, among objects) within the same image. This strategy allows us to remove the bottleneck caused by the limited perceptive fields of CNNs encoders, leading to naturally finer-grained predictions, as shown in Fig.~\hyperref[fig:ds]{1}.
We evaluate the performance of MonoViT on the popular KITTI dataset, using the standard split by Eigen \textit{et al}.~\cite{Eigen2014}. The comparison to SoTA solutions highlights the constantly superior accuracy of our framework. Moreover, we also analyze the generalization capability of self-supervised monocular depth estimation networks across different datasets. Purposely, we compare MonoViT with its main competitors on the Make3D~\cite{saxena2008make3d} and DrivingStereo~\cite{yang2019drivingstereo} datasets, highlighting, even in this case, the superior generalization capacity of MonoViT.

\input{3relatedwork}

\input{4method}

\input{5experiment}
\input{6conclusion}


{\small
\bibliographystyle{ieee_fullname}
\bibliography{egbib}
}

\end{document}

%% file: 3relatedwork.tex
\section{Related works}

This section reviews the literature concerning self-supervised monocular depth estimation and ViT architectures, being both relevant to our work.


\textbf{Monocular Depth Estimation.} Estimating depth from a single image is a challenging, inherently ill-posed problem. Nonetheless, the many learning-based approaches aimed at addressing it \cite{zhao2020monocular} enabled significant progress in the field. As fully supervised techniques \cite{laina2016, Liu_IEEE_2016, Fu2018,bhat2021adabins} for depth estimation advanced rapidly, the availability of precise depth labels in the real world became a major issue. Hence, more recent self-supervised works provided alternatives to remove it, avoiding the need for hard-to-source ground-truth depth annotations. This goal is feasible by casting the depth estimation task as a view-synthesis problem between adjacent views, in space or time, of the same observed scene. Precisely, training single-view depth estimation network with stereo images \cite{garg16,Godard17}, monocular videos \cite{zhou_sfmlearner}, or a combination of both \cite{Zhan_CVPR_2018, monodepth2}.  
The supervisory signal based on the photometric difference between real and synthesized images enables training in a self-supervised manner.

\begin{figure*}[t]
\begin{center}
 \includegraphics[width=0.8\linewidth]{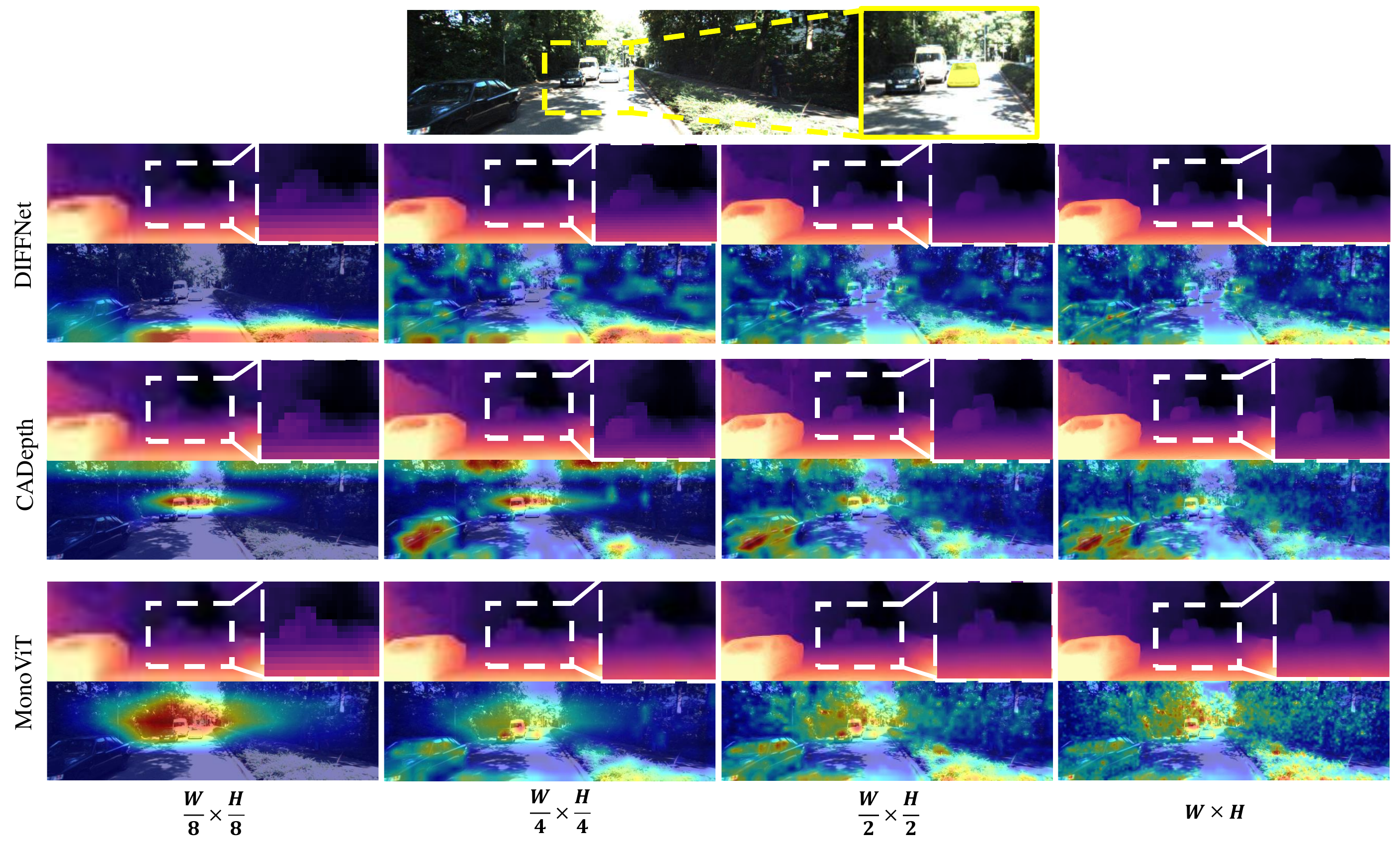}
\end{center}
   \caption{\textbf{Attention maps of SoTA methods and our MonoViT.}
   The first row shows the RGB image, and the highlighted car is the region we want to analyze. In the next two rows, we report multiscale disparity predictions and attention maps of each method. For an object that is small in size and hard to distinguish from the background, such as the car highlighted, we notice how MonoViT can predict its disparity even at the lowest resolution (\emph{i.e.} $\frac{H}{8} \times \frac{W}{8}$). At the same time, other methods fail to capture it. 
   }
\label{fig:atten}
\end{figure*}

Although stereo pairs enable scale recovery, with further improvements achievable by leveraging noisy proxy labels \cite{Tosi_CVPR_2019, Watson_ICCV_2019, Choi_2021_ICCV}, guidance from visual odometry \cite{vomonodepth19} or trinocular assumptions \cite{Poggi_3DV_2018}, unlabeled video sequences represent a more flexible alternative at the expense of learning camera poses alongside depth. Several frameworks have advanced this line of research by incorporating additional losses and constraints such as those based on  direct visual odometry \cite{Wang_CVPR_2018}, adversarial learning\cite{zhao2020masked}, ICP \cite{Mahjourian_CVPR_2018},  normal consistency \cite{Yang_CVPR_2018, Yang_2018}, semantic segmentation \cite{kumar2021syndistnet, guizilini2020semantically} and uncertainty \cite{poggi2020uncertainty, yang2020d3vo}. Another notable example is given in~\cite{monodepth2} where the authors introduced a minimum reprojection loss between frames and an auto-masking strategy to handle both occluded regions and static camera situations that violate the main constraints of the view-synthesis formulation and, as a consequence, cause poor network convergence. Other works, instead, directly tackle highly complex scenarios \cite{zhao2021unsupervised} and model rigid and non-rigid components present in the scene using the estimated depth, relative camera poses, and optical flow in order to handle independent motions \cite{tosi2020distilled, yin2018geonet, Zou18, Chen_ICCV_2019, ranjan2019, Yang_2018_ECCV_Workshops} or by means of scene decomposition \cite{safadoust2021self}.

\textbf{Network architectures.} Playing with different architectures used as backbone showed a significant impact on the performance of monocular depth estimation itself. Yin \textit{et al.}~\cite{yin2018geonet} replaced the VGG encoder used by~\cite{zhou_sfmlearner} with a ResNet. Guizilini \textit{et al.}~\cite{guizilini2020} designed a novel model, PackNet, to learn
detail-preserving compression and decompression of features by using 3D convolutions. Lyu \textit{et al.}~\cite{hrnet2020} worked on the features decoding, implementing an attention module for multi-scale feature fusion. Because of the limited receptive field of CNNs, the network performance still has room for further improvement. To extract the long-range relationships between features, Yan \textit{et al.}~\cite{yan2021channel} propose a channel-wise attention-based network to aggregate discriminated features in channel dimensions. Considering the ability of HRNet~\cite{hrnet2020} at modeling multi-scale features, Zhou \textit{et al.}~\cite{zhoudiffnet} introduced HRNet for self-supervised monocular depth estimation. Other works, instead, focused on the design of efficient and fast networks suitable for low-powered embedded devices \cite{Poggi_IROS_2018, DATE_2019, icra_2019_fastdepth}. Despite the increased accuracy achieved by the above networks, the issue concerning long-range relationships persists~\cite{li2022depthformer}.

\begin{figure*}[ht]
\begin{center}
 \includegraphics[width=0.8\linewidth]{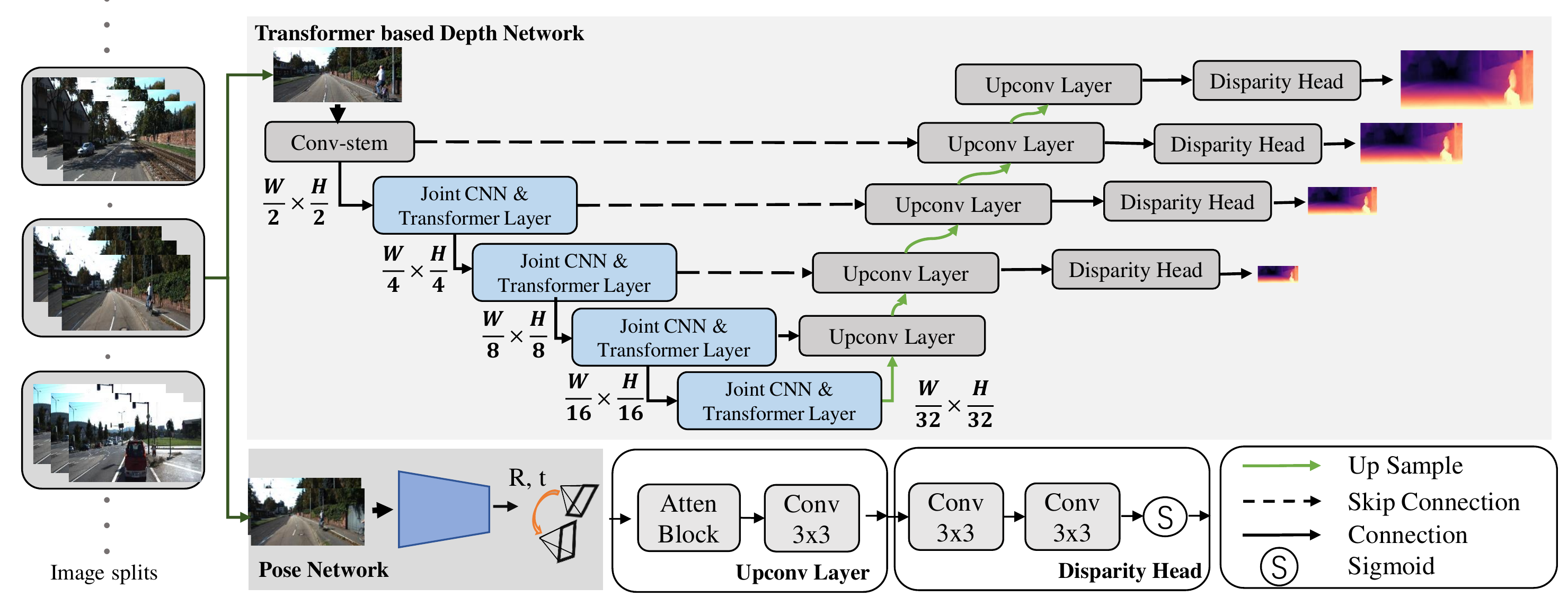}
\end{center}
\vspace{-0.2cm}
\caption{\textbf{Overview of our MonoViT architecture}. Our MonoViT consists of two parts, Depth Network and Pose Network. For Depth Network, both Transformer~\cite{lee2021mpvit} and convolutional layer are adopted to enhance the feature modeling and depth inferring. For pose estimation between temporally adjacent images, we use a lightweight PoseNet as in previous works~\cite{monodepth2,lyu2020hr,zhoudiffnet,yan2021channel}.
}
 
\label{fig:arch}
\end{figure*}

\textbf{Transformers in Depth Estimation.}
Recently, inspired by the success of the attention mechanism on modeling global context perception, ViTs~\cite{dosovitskiy2020image,liu2021swin} showed great potential in tasks such as image classification~\cite{liu2021swin,lee2021mpvit}, object detection~\cite{dai2021up,carion2020end}, and semantic segmentation~\cite{wang2021max,xie2021segformer}. A few works also tackled monocular depth estimation by using Transformer architectures~\cite{ranftl2021vision,yang2021transformer,li2022depthformer}. However, these methods focus on the supervised setting only.

%% file: 4method.tex
\section{Proposed framework}

This section analyzes the necessity for introducing a Transformer for self-supervised monocular depth estimation. Then, we describe our MonoViT network architecture and the loss functions used for the self-supervised training of our framework.

\subsection{Motivations}

\begin{figure}[t]
\begin{center}
 \includegraphics[width=0.9\linewidth]{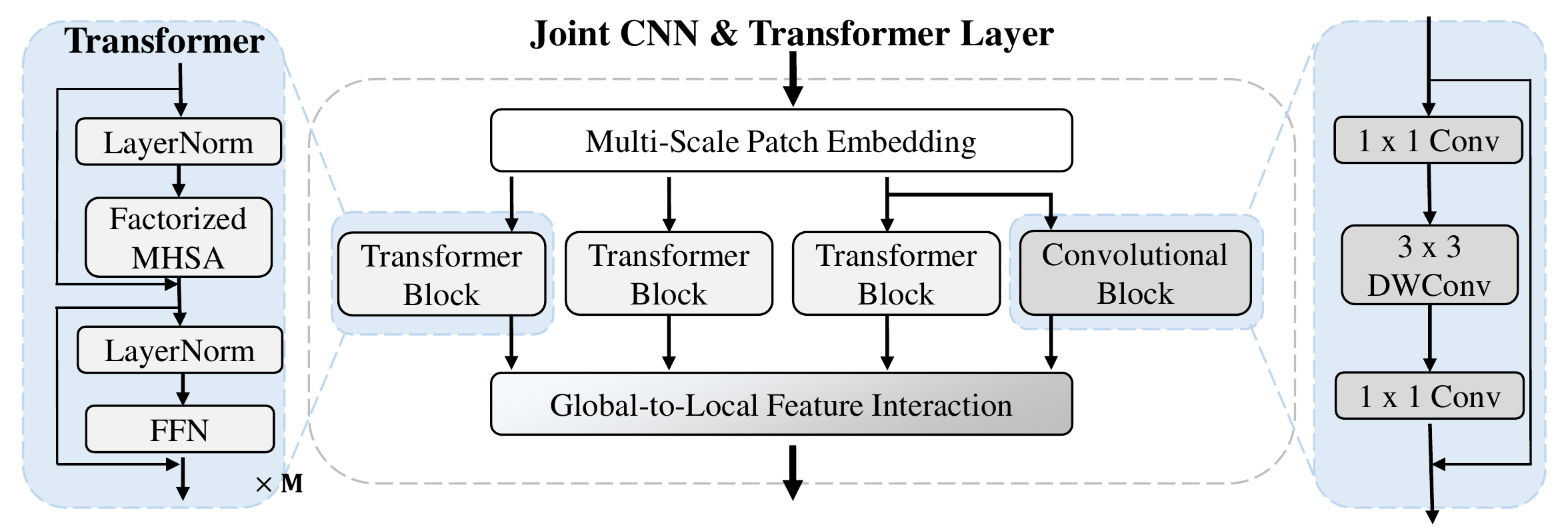}
\end{center}
\vspace{-0.2cm}
   \caption{
   \textbf{Joint CNN \& Transformer Layer used in depth encoder}. Each Transformer block contains $M$ Transformer layers, consisting of a Layer Normalization (LayerNorm) module, a Factorized Multi-head Self Attention (MHSA) layer~\cite{lee2021mpvit}, another Layer Normalization and a Feed-forward Network (FFN).}
\label{fig:vit}
\end{figure}

Unlike supervised depth estimation methods, the supervisory signal of self-supervised approaches derives from image reprojection across different, nearby viewpoints. Thus, to achieve good performance, this formulation requires the network to accurately perceive the scene structure: a challenging task, especially for regions with hard to distinguish foreground objects from the background. Current SoTA networks~\cite{yan2021channel,lyu2020hr} rely on traditional convolutional layers for aggregating context information and gradually lift the receptive field of the network through a cascade of layers and strided convolution~\cite{Unet}. However, given the intrinsic locality of the convolution operator, CNNs hardly model long-range appearance similarity among objects, in particular within the shallowest features. An example of this occurs when the foreground objects have a texture similar to the one of the background. In such a case, the feature backbone tends to embed them in the same semantic context, and the whole architecture cannot distinguish between foreground and background depths. Fig.~\ref{fig:atten} shows this behaviour; we can notice that the car in the middle of the road is hard to spot from the ground due to the strong sunlight. CNNs such as CADepth~\cite{yan2021channel} and DIFFNet~\cite{zhoudiffnet} predict a depth for the car similar to the one of the ground plane. This fact is due to their encoder network paying more attention to the ground than the car itself. Hence, we propose integrating convolutions and ViT blocks to address the standard limitation of the former, since the latter has more significant potential for modeling long-range correlation.

Driven by this rationale, we design our Monocular Vision Transformer framework, \textbf{MonoViT} in short, as shown in Fig.~\ref{fig:arch}.
It includes a DepthNet and a PoseNet, respectively, designed for depth prediction of each input image and pose estimation and trained through image reconstruction losses.

\subsection{DepthNet Architecture}

As typical in previous works~\cite{zhou_sfmlearner,monodepth2}, we design our DepthNet as an encoder-decoder architecture.

\textbf{Depth encoder.} As pointed out by recent research~\cite{hrnet2020,yan2021channel,zhou_sfmlearner,monodepth2}, the encoder is crucial for effective features extraction. Inspired by one of the most recent Transformer architectures -- \textit{i.e.}, MPViT~\cite{lee2021mpvit}, in which a Multi-Path Transformer Block is proposed for simultaneously representing local and global context extracted from images -- we follow such a design to build the key components of our depth encoder in five stages. Given the current input image, we adopt a Conv-stem block consisting of two convolutions with kernel size $3 \times 3$ and stride of 2 only at the first convolution, generating features with size $\frac{H}{2} \times \frac{W}{2}$. From stage two to stage five, we stack the Multi-Path Transformer Blocks in each stage, shown as ``Joint CNN \& Transformer Layer'' in Fig.~\ref{fig:arch}. Precisely, each ``Joint CNN \& Transformer Layer'' (shown in Fig.~\ref{fig:vit}) consists of a Multi-Scale Patch Embedding layer, used to embed various-sized visual tokens in parallel -- in our case, four parallel convolutional blocks extract features with a receptive field of $3\times3$,  $3\times3$, $5\times5$ and $7\times7$ pixels by stacking multiple $3\times3$ convolutional layers. Then, considering the advantage of ViT at building global dependencies while shows limitations modeling local details~\cite{lee2021mpvit}, extracted tokens are processed through both convolutional layers and Transformers blocks, in a parallel and complementary manner -- \emph{i.e.}, using the four branches shown in Fig.~\ref{fig:vit}, respectively three parallel Transformer blocks and a convolutional block, this latter made of 1$\times$1, 3$\times$3 depthwise and 1$\times$1 convolutions. While the convolutional branch constructs the local relationship between neighbors within features $\mathbf{L}$, the three Transformer Blocks model the information interaction over the whole input space within features $\mathbf{G}_0, \mathbf{G}_1, \mathbf{G}_2$, thanks to the self-attention mechanism. 
Specifically, these latter take a sequence of visual tokens embedded by the Multi-Scale Patching Embedding module and project them into a query ($\mathbf{Q}$), key ($\mathbf{K}$), and value ($\mathbf{V}\in \mathbb{R}^{N\times C}$) vectors through three separated but structure same heads (where $N$ denotes the number of visual tokens, equal to the total number of pixels in the input space). The self-attention mechanism is implemented in an efficient factorized way~\cite{lee2021mpvit}:
\begin{equation}
    \text{FactorAtt}(\mathbf{Q}, \mathbf{K}, \mathbf{V}) = \frac{\mathbf{Q}}{\sqrt{C}}(\text{softmax}(\mathbf{K})^{T}\mathbf{V}),
\end{equation}
where $C$ refers to the embedding dimension.
Finally, a feature fusion block is used to collect and further enhance the interaction between local and global features extracted by the ``Joint CNN \& Transformer Layer'' at stage $i$.

\begin{equation}
    \mathbf{A}_{i} = \texttt{Concat}([\mathbf{L}_i, \mathbf{G}_{i,0}, \mathbf{G}_{i,1}, \mathbf{G}_{i,2}]),
\end{equation}
\begin{equation}
    \mathbf{X}_{i+1} = \mathcal{H}(\mathbf{A}_{i}),
\end{equation}
with $\mathbf{A}_i\in \mathbb{R}^{H_i \times W_i \times C_i}$ being the aggregated feature and $\mathcal{H}(\cdot)$ a 1$\times$1 convolutional layer which fuses them and yields the final feature $\mathbf{X}_{i+1}$ for the next stage $i+1$.

A clear benefit of integrating convolutions with Transformer is the comprehensive -- both local and global -- interaction between pixels. It helps the network to perceive the structure and relative position of objects so that small foreground objects can be preserved even at the lowest resolution, rather than collapsed into similar texture background as shown in Fig. \ref{fig:atten}.

\textbf{Depth decoder.} 
Taking the multi-scale features from the depth encoder, cross-layer and cross-scale connections are adopted in our depth decoder to gradually increase the spatial resolution, as shown in Fig.~\ref{fig:arch}. Considering the context difference between features at different scales, e.g. higher resolution features favour fine-grained details, we enhance cross-scale feature fusion with both spatial and channel attention mechanisms~\cite{lyu2020hr,zhoudiffnet} (\emph{i.e.}, our Atten Block). Finally, four heads -- made of two convolutional layers and a Sigmoid activation -- are in charge of disparity (inverse depth) prediction from corresponding aggregated features, outputting maps at full, $\frac{1}{2}$, $\frac{1}{4}$, $\frac{1}{8}$ resolution respectively.

\subsection{PoseNet}
Following~\cite{monodepth2,lyu2020hr,zhoudiffnet,yan2021channel}, our PoseNet favors a simple, yet effective implementation. Specifically, our PoseNet uses the lightweight structure of ResNet18~\cite{resnet}. Receiving concatenated images [$\mathcal{I}$, $\mathcal{I}$\textsuperscript{\textdagger}] as input, it outputs a 6 DoF relative pose $\mathbf{T}$ between adjacent frames of a video sequence.

\begin{table*}[ht]
\centering
\scalebox{0.62}{
\renewcommand{\tabcolsep}{2pt}
\begin{tabular}{ccc}

\begin{tabular}{|l |c|c||c c c c| c c c|}
\hline
& & &\multicolumn{4}{|c|}{lower is better} & \multicolumn{3}{|c|}{higher is better}\\
Method & Data & Resolution &Abs Rel& Sq Rel&  RMSE & RMSE log & $\delta_{1}$ & $\delta_{2}$ & $\delta_{3} $  \\
\hline
Monodepth2~\cite{monodepth2}     & M    & 640$\times$192 & 0.115 & 0.903 & 4.863 & 0.193 & 0.877 & 0.959 & 0.981\\
Sun~\cite{sun2021unsupervised}   & M    & 640$\times$192 & 0.117 & 0.863 & 4.813 & 0.192 & 0.871 & 0.959 & 0.982\\
SGDepth~\cite{klingner2020self}  & M+Se & 640$\times$192 & 0.113 & 0.835 & 4.693 & 0.191 & 0.879 & 0.961 & 0.981\\
SAFENet~\cite{choi2020safenet}   & M+Se & 640$\times$192 & 0.112 & 0.788 & 4.582 & 0.187 & 0.878 & 0.963 & 0.983\\
VC-Depth~\cite{zhou2020constant} & M    & 640$\times$192 & 0.112 & 0.816 & 4.715 & 0.190 & 0.880 & 0.960 & 0.982\\
PackNet$^{\ddagger}$~\cite{guizilini2020}     & M    & 640$\times$192 &0.108 &0.727 &4.426 &0.184 &0.885 &0.963 &0.983\\
Mono-Uncertainty\cite{poggi2020uncertainty}
                                 & M    & 640$\times$192 & 0.111 & 0.863 & 4.756 & 0.188 & 0.881 & 0.961 & 0.982\\
HR-Depth~\cite{lyu2020hr}        & M    & 640$\times$192 & 0.109 & 0.792 & 4.632 & 0.185 & 0.884 & 0.962 & 0.983\\
Johnston et al.~\cite{johnston2020self} & M    & 640$\times$192 & 0.106 & 0.861 & 4.699 & 0.185 & 0.889 & 0.962 & 0.982\\
CADepth$^{\star}$~\cite{yan2021channel}    & M    & 640$\times$192 & 0.105 & 0.769 & 4.535 & 0.181 & 0.892 & 0.964 & 0.983\\
DIFFNet$^{\dagger}$~\cite{zhoudiffnet}       & M    & 640$\times$192 & 0.102 & 0.749 & 4.445 & 0.179 & 0.897 & 0.965 & 0.983\\
MonoFormer~\cite{bae2022monoformer} & M & 640$\times$192 & 0.106 & 0.839 & 4.627 & 0.183 & 0.889 & 0.962 & 0.983 \\
\hline
\bf{MonoViT (ours)}       & M    & 640$\times$192 & \bf{0.099} & \bf{0.708} & \bf{4.372} & \bf{0.175} & \bf{0.900} & \bf{0.967} & \bf{0.984}\\
\hline
\multicolumn{10}{c}{}\\
\multicolumn{10}{c}{}\\
\hline
Monodepth2~\cite{monodepth2} & MS & 640$\times$192 & 0.106 & 0.818 & 4.750 & 0.196 & 0.874 & 0.957 & 0.979\\
HR-depth~\cite{lyu2020hr}    & MS & 640$\times$192 & 0.107 & 0.785 & 4.612 & 0.185 & 0.887 & 0.962 & 0.982\\
CADepth$^{\star}$~\cite{yan2021channel}& MS & 640$\times$192 & 0.102 & 0.752 & 4.504 & 0.181 & 0.894 & 0.964 & 0.983\\
DIFFNet$^{\dagger}$~\cite{zhoudiffnet}   & MS & 640$\times$192 & 0.101 & 0.749 & 4.445 & 0.179 & 0.898 & 0.965 & 0.983\\
\hline
\bf{MonoViT (ours)}       & MS    & 640$\times$192 & \bf{0.098} & \bf{0.683} & \bf{4.333} & \bf{0.174} & \bf{0.904} & \bf{0.967} & \bf{0.984}\\
\hline
\end{tabular}

& \quad &

\begin{tabular}{|l |c|c||c c c c| c c c|}
\hline
Method & Data &Resolution &\multicolumn{4}{|c|}{lower is better} & \multicolumn{3}{|c|}{higher is better}\\
~ & ~ & ~ &Abs Rel& Sq Rel&  RMSE & RMSE log & $\delta_{1}$ & $\delta_{2}$ & $\delta_{3} $  \\
\hline
Monodepth2~\cite{monodepth2}    & M & 1024$\times$320 & 0.115 & 0.882 & 4.701 & 0.190 & 0.879 & 0.961 & 0.982\\
Sun~\cite{sun2021unsupervised}  & M & 1024$\times$320 & 0.110 & 0.791 & 4.557 & 0.184 & 0.887 & 0.964 & 0.983 \\
SAFENet~\cite{choi2020safenet}  & M+Se & 1024$\times$320 & 0.106 & 0.743 & 4.489 & 0.181 & 0.884 & 0.965 & \bf{0.984} \\
HR-Depth~\cite{lyu2020hr}       & M & 1024$\times$320 & 0.106 & 0.755 & 4.472 & 0.181 & 0.892 & 0.966 & \bf{0.984}\\
FeatDepth~\cite{shu2020feature} & M & 1024$\times$320 & 0.104 & 0.729 & 4.481 & 0.179 & 0.893 & 0.965 & \bf{0.984}\\
GCNDepth~\cite{masoumian2021gcndepth}
                                & M & 1024$\times$320 & 0.104 & 0.720 & 4.494 & 0.181 & 0.888 & 0.965 & \bf{0.984}\\
CADepth$^{\star}$~\cite{yan2021channel}   & M & 1024$\times$320 & 0.102 & 0.734 & 4.407 & 0.178 & 0.898 & 0.966 & \bf{0.984} \\
DIFFNet$^{\dagger}$~\cite{zhoudiffnet}      & M & 1024$\times$320 & 0.097 & 0.722 & 4.345 & 0.174 & 0.907 & 0.967 & \bf{0.984} \\
\hline
\bf{MonoViT (ours)}       & M    & 1024$\times$320 & \bf{0.096} & \bf{0.714} & \bf{4.292} & \bf{0.172} & \bf{0.908} & \bf{0.968} & \bf{0.984}\\
\hline
PackNet$^{\ddagger}$~\cite{guizilini2020}    & M & 1280$\times$384 &0.104 &0.758 &4.386 &0.182 &0.895 &0.964 &0.982\\
SGDepth~\cite{klingner2020self} & M+Se & 1280$\times$384 & 0.107 & 0.768 & 4.468 & 0.186 & 0.891 & 0.963 & 0.982\\
HR-Depth~\cite{lyu2020hr}       & M   & 1280$\times$384 & 0.104 & 0.727 & 4.410 & 0.179 & 0.894 & 0.966 & \bf{0.984}\\
CADepth$^{\star}$~\cite{yan2021channel} & M & 1280$\times$384 & 0.102 & 0.715 & 4.312 & 0.176 & 0.900 & 0.968 & \bf{0.984} \\
\hline
\bf{MonoViT (ours)}                       & M & 1280$\times$384 & \bf{0.094} & \bf{0.682} & \bf{4.200 }& \bf{0.170} & \bf{0.912} & \bf{0.969} & \bf{0.984}\\
\hline
\multicolumn{10}{c}{}\\
\hline
Monodepth2~\cite{monodepth2}    & MS & 1024$\times$320 & 0.106 & 0.818 & 4.750 & 0.196 & 0.874 & 0.957 & 0.979\\
HR-Depth~\cite{lyu2020hr}       & MS & 1024$\times$320 & 0.101 & 0.716 & 4.395 & 0.179 & 0.892 & 0.966 & 0.984\\
CADepth$^{\star}$~\cite{yan2021channel} & MS & 1024$\times$320 & 0.096 & 0.694 & 4.264 & 0.173 & 0.908 & 0.968 & 0.984 \\
DIFFNet$^{\dagger}$~\cite{zhoudiffnet}  & MS & 1024$\times$320 & 0.094 & 0.678 & 4.250 & 0.172 & 0.911 & 0.968 & 0.984 \\
\hline
\bf{MonoViT (ours)}                       & MS & 1024$\times$320 & \bf{0.093} & \bf{0.671} & \bf{4.202} & \bf{0.169} & \bf{0.912} & \bf{0.969} & \bf{0.985}\\
\hline
\end{tabular}
\end{tabular}
}
\caption{\textbf{Results on the KITTI benchmark using the Eigen split~\cite{kitti}}. Each method is grouped by input resolution (low: left, high: right) and training methodology (M: monocular videos, MS: binocular videos, Se: trained with semantic labels). The best scores are in {\bf bold}. $\star$ refers to the current SoTA self-supervised method on the KITTI depth benchmark. 
$\dagger$ stands for the novel results from the official Github repository, better than published ones.
${\ddagger}$ refers to the model pretrained on Cityscapes~\cite{cityscapes}, while the others are pretrained on ImageNet~\cite{imagenet}.}
\label{tab:k}
\end{table*}

\subsection{Self-supervised Learning}

We cast depth estimation as an image reconstruction task, replacing ground truth labels with unlabeled, monocular videos at training time.
The depth network takes a still (\ie, target) image $\mathcal{I}$ and predict its dense inverse depth map $d$, from which depth $\mathcal{D}$ is derived as $1./d$ and forcing it to be in $[\mathcal{D}_{min}, \mathcal{D}_{max}]$ as in~\cite{monodepth2}.

\textbf{View reconstruction loss.} 
By knowing camera intrinsics $\mathbf{k}$ and the predicted pose $\mathbf{T}$ between two nearby views, a reconstructed target image $\tilde{\mathcal{I}}$ is obtained as a function $\pi$ of intrinsics, pose, source image $\mathcal{I}$\textsuperscript{\textdagger} and depth $\mathcal{D}$. A loss signal $\mathcal{L}_{ss}$ is computed as a function $\mathcal{F}$ of inputs $\tilde{\mathcal{I}}$ and $\mathcal{I}$:

\begin{equation}
    \mathcal{L}_{ss} = \mathcal{F}(\tilde{\mathcal{I}}, \mathcal{I}) = \mathcal{F}( \pi(\mathcal{I}\textsuperscript{\textdagger}, 
\mathbf{T}, \mathbf{k}, \mathcal{D}), \mathcal{I}).
\label{eq:reproj}
\end{equation}
$\mathcal{F}$ is usually obtained as a weighted sum between a structural similarity term and an intensity difference term. Popular choices for these two terms are the Structured Similarity Index Measure (SSIM) \cite{SSIM} and the L1 difference, as proposed in~\cite{monodepth2}:

\begin{equation}
    \mathcal{F}(\tilde{\mathcal{I}}, \mathcal{I}) = \alpha \cdot \frac{1- \text{SSIM}(\tilde{\mathcal{I}}, \mathcal{I})}{2} + (1-\alpha) \cdot|\tilde{\mathcal{I}}-\mathcal{I}|
\end{equation}
with $\alpha$ commonly set to 0.85~\cite{monodepth2}. Besides, for each pixel $p$, the minimum among losses computed from forward and backward adjacent frames allows for softening the effect of occlusions~\cite{monodepth2} on the reprojection process

\begin{equation}
    \mathcal{L}_{ss}(p) = \min_{i \in [1,-1]} \mathcal{F}(\tilde{\mathcal{I}}_i(p), \mathcal{I}(p))
\end{equation}
with `1' and `-1' referring to the forward and backward adjacent frames, respectively.

\textbf{Smoothness loss.}
 As in previous works~\cite{monodepth2,zhoudiffnet}, the edge-aware smoothness loss is used to improve the inverse depth map $d$:
\begin{equation}
\mathcal{L}_{smooth} = |\partial_{x} d^{*}|e^{\partial_{x} I} + |\partial_{y} d^{*}|e^{\partial_{y} I}, \label{eq:smooth}
\end{equation}
where $d^{*}=d/\hat{d}$ represents the mean-normalized inverse depth. Besides, following~\cite{monodepth2}, an auto-mask $\mu$ is calculated to filter static frames and objects moving with the same motion of the camera.

\textbf{Total loss.} Finally, both the view reconstruction loss $\mathcal{L}_{ss}$ and the smoothness loss $\mathcal{L}_{smooth}$ are computed from outputs at each scale $s \in \{1, \frac{1}{2}, \frac{1}{4}, \frac{1}{8} \}$ -- brought to full resolution -- and then averaged as $\mathcal{L}_{tot}$ to train MonoViT:
\begin{equation}
\mathcal{L}_{tot} = \frac{1}{4} \cdot \sum_{s=1}^{4} (\mu \cdot \mathcal{L}_{ss} + \lambda \cdot \mathcal{L}_{smooth}), \label{eq:total}
\end{equation}
with $\lambda$ being set to $10^{-3}$.

%% file: 5experiment.tex
\begin{table}[t]
\centering
\scalebox{0.7}{
\renewcommand{\tabcolsep}{2pt}
\begin{tabular}{|l |c||c c c c| c c c|}
\hline
& &\multicolumn{4}{|c|}{lower is better} & \multicolumn{3}{|c|}{higher is better}\\
Method & Data   &Abs Rel& Sq Rel&  RMSE & RMSE log & $\delta_{1}$ & $\delta_{2}$ & $\delta_{3} $  \\
\hline
Monodepth2~\cite{monodepth2}     & M  & 0.090 & 0.545 & 3.942 & 0.137 & 0.914 & 0.983 & 0.995\\
Johnston~\cite{johnston2020self} & M  & 0.081 & 0.484 & 3.716 & 0.126 & 0.927 & 0.985 & 0.996\\
HR-Depth~\cite{lyu2020hr}        & M  & 0.085 & 0.471 & 3.769 & 0.130 & 0.919 & 0.985 & 0.996\\
CADepth~\cite{yan2021channel}    & M  & 0.080 & 0.450 & 3.649 & 0.124 & 0.927 &  0.986 & 0.996  \\
DIFFNet~\cite{zhoudiffnet}       & M  & 0.076 & 0.412 & 3.494 & 0.119 & 0.935 & 0.988 & 0.996 \\
\bf{MonoViT (ours)}                        & M  & \bf{0.075} & \bf{0.389} & \bf{3.419} & \bf{0.115} & \bf{0.938} & \bf{0.989} & \bf{0.997} \\
\hline
\multicolumn{9}{c}{ } \\
\hline
PackNet$^{\ddagger}$~\cite{guizilini2020}     & M  & 0.071 & 0.359 & 3.153 & 0.109 & 0.944 & 0.990 & 0.997\\
\bf{MonoViT (ours)}                        & M & \bf{0.067} & \bf{0.328} & \bf{3.108} & \bf{0.104} & \bf{0.950} & \bf{0.992} & \bf{0.998} \\
\hline
\end{tabular}
}
\caption{\textbf{Results on KITTI using the improved ground truth~\cite{Uhrig2017THREEDV}.} Top: $640\times192$ input resolution, bottom: $1280\times384$. ${\ddagger}$ pretrained on Cityscapes~\cite{cityscapes} (on ImageNet~\cite{imagenet} otherwise).}
\label{tab:kigt}
\end{table}

\section{Experiments}

In this section, we report the outcome of our experiments, clearly supporting the superior accuracy of MonoViT at estimating depth across several benchmarks.

\subsection{Implementation Details}
We implement our MonoViT in Pytorch. The model is trained for 20 epochs on the KITTI dataset using AdamW~\cite{adamw} as optimizer and a batch size set to 12. The initial learning rate for PoseNet and depth decoder is 10$^{-4}$, while the Transformer-based depth encoder is trained with an initial learning rate of 5$\times10^{-5}$. The number $M$ of Transformer layers in each of the three Transformer blocks in the `Joint CNN \& Transformer Layer' is set as 1, 3, 6, 3 from stage 2 to stage 5 in the depth encoder, respectively.
Both the pose encoder and depth encoder are pre-trained on ImageNet~\cite{imagenet}. We use a single RTX 3090 GPU for the low resolution ($640\times192$) experiments while 4 RTX 3090 GPUs for higher resolution ($1024\times320$,$1280\times384$) ones. Overall, network training  requires about 15 hours. In our experiments, we adopt the same data augmentation detailed in~\cite{monodepth2,lyu2020hr}.

For evaluation, we compute the seven standard metrics (Abs Rel, Sq Rel, RMSE, RMSE log, $\delta_{1} < 1.25$, $\delta_{2} < 1.25^2$, $\delta_{3} < 1.25^3$) proposed in~\cite{Eigen2014} and used by most works in the literature.

\begin{figure*}[t]
\begin{center}
 \includegraphics[width=0.9\linewidth]{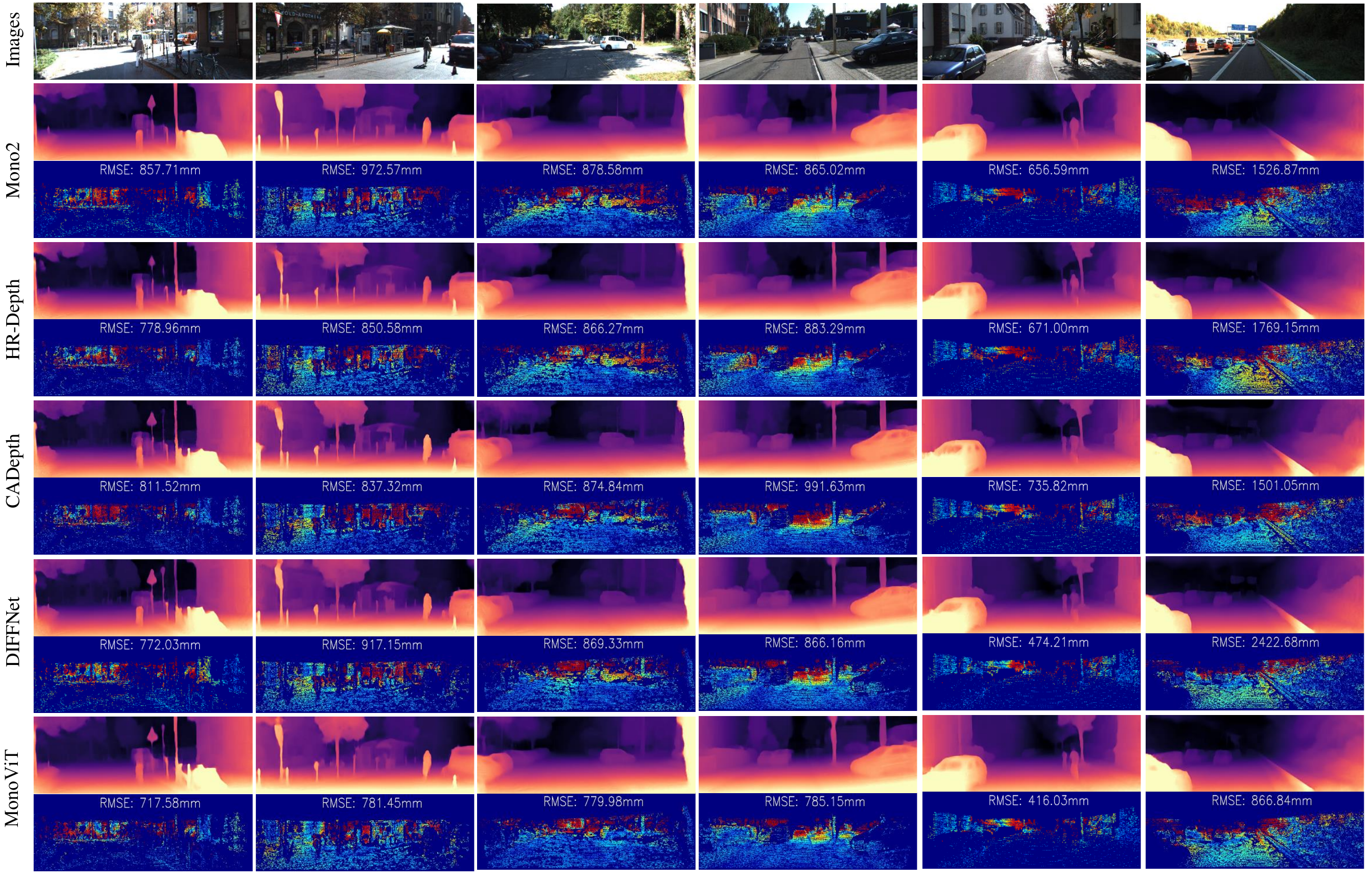}
\end{center}
    \vspace{-0.4cm}
   \caption{\textbf{Qualitative results on KITTI.} Top row, input images. Then, predictions by SoTA methods (Mono2~\cite{monodepth2}, HR-Depth~\cite{lyu2020hr}, CADepth~\cite{yan2021channel}, DIFFNet~\cite{zhoudiffnet}) and MonoViT (Ours). For each method, we show the depth map and the corresponding error map.}
   \vspace{-0.4cm}
\label{fig:vk}
\end{figure*}

\subsection{Datasets}

\textbf{KITTI~\cite{kitti}.} The KITTI stereo dataset contains 61 scenes, with a typical image size of $1242 \times 375$, captured using a stereo rig mounted
on a moving car equipped with a LiDAR sensor. Following previous works in this field~\cite{monodepth2,zhoudiffnet,lyu2020hr}, we use the image split of Eigen \textit{et al.}~\cite{Eigen2014}, which consists of 39810 monocular triplets for training and 4424 for validation. To compare with the existing solutions, we evaluate the single-view depth performance on the test split of \cite{Eigen2014} either using raw LiDAR (697 images) or improved ground truth labels~\cite{Uhrig2017THREEDV} (652 images).

\textbf{Make3D~\cite{saxena2008make3d}.} This dataset features outdoor environments and is typically used for testing the generalization performance of monocular depth frameworks. We test MonoViT following the same image pre-processing steps and computing the evaluation metrics detailed in~\cite{monodepth2}.

\textbf{DrivingStereo~\cite{yang2019drivingstereo}.} It is a large-scale stereo dataset depicting autonomous driving scenarios. Among several sequences, we use the four image splits made available on the website, each made of 500 frames collected under different weather conditions, respectively \textit{foggy}, \textit{cloudy}, \textit{rainy} and \textit{sunny}. We use this dataset to evaluate the generalization capacity of MonoViT and its most recent competitors.

\subsection{Depth Evaluation}

\textbf{Results on KITTI:}
We test our model by using the standard KITTI Eigen split~\cite{Eigen2014}, which includes 697 images coupled with raw LiDAR scans. Among them, improved ground truth labels~\cite{Uhrig2017THREEDV} are provided for 652 images. Since monocular depth models trained on video sequences suffer from monocular scale ambiguity, the estimated depth is scaled by the per-image median ground truth~\cite{zhou_sfmlearner}. 

Tab.~\ref{tab:k} collects the results achieved by SoTA self-supervised frameworks, processing either low resolution (left) or high resolution (right) images. We report results for methods trained both using monocular (`M', top), and binocular videos (`MS', bottom) for completeness. MonoViT significantly outperforms existing SoTA methods for any training resolution and setting on all metrics. In particular, we also highlight how MonoViT greatly outperforms MonoFormer~\cite{bae2022monoformer}, a concurrent attempt to deploy Transformers in self-supervised monocular depth estimation.
Tab.~\ref{tab:kigt} reports the same metrics computed over the improved ground truth labels processing $640\times192$ images. Again, MonoViT is constantly more accurate.

Fig.~\ref{fig:vk} reports a comparison between MonoViT and some of its competitors, showing that our model can get a much lower RMSE and proving that MonoViT is more powerful at modeling relations between objects than existing models.

\begin{table}[t]
\centering
\setlength
\tabcolsep{4.2pt}{
\footnotesize
\begin{tabular}{|l|| c c c c |}
\hline
 &\multicolumn{4}{|c|}{lower is better} \\
Method &Abs Rel& Sq Rel&  RMSE & RMSE log  \\
\hline
Monodepth2~\cite{monodepth2}&  0.321 & 3.378 &  7.252 &  0.163  \\

HR-Depth~\cite{lyu2020hr}  & 0.305 &2.944 &6.857 & 0.157\\
CADepth~\cite{yan2021channel} & 0.319 & 3.564 & 7.152 &  0.158 \\
DIFFNet~\cite{zhoudiffnet} &0.298 &2.901 &6.753 & 0.153  \\
\hline
\bf{MonoViT (ours)} & \bf{0.286} & \bf{2.758} & \bf{6.623} & \bf{0.147}  \\
\hline
\end{tabular}}
\caption{\textbf{Quantitative results on the Make3D Dataset~\cite{saxena2008make3d}.} Models trained on KITTI with $640\times192$ images.}
\vspace{-0.4cm}
\label{tab:make3d}
\end{table}



\textbf{Results on Make3D.}
We run experiments on the Make3D dataset~\cite{saxena2008make3d} in order to evaluate the capability of our model to generalize on different real-world environments. By following the same protocol indicated in~\cite{monodepth2,lyu2020hr,zhoudiffnet,yan2021channel}, we firstly train our model on KITTI using images at $640\times192$ resolution and, then test on Make3D without a fine-tuning procedure.
For fairness, we evaluate MonoViT and the existing self-supervised networks using the same evaluation code provided by \cite{monodepth2}. Tab. \ref{tab:make3d} demonstrates how our proposed architecture allows us to outperform other strategies by a large margin and to achieve SoTA generalization results. 

\begin{table}[t]
\centering
\scalebox{0.65}{
\renewcommand{\tabcolsep}{2pt}
\begin{tabular}{|c|l||c c c c| c c c|}
\hline
& &\multicolumn{4}{|c|}{lower is better} & \multicolumn{3}{|c|}{higher is better}\\
Domain & Method & Abs Rel& Sq Rel&  RMSE & RMSE log & $\delta_{1}$ & $\delta_{2}$ & $\delta_{3} $  \\
\hline
~& Monodepth2~\cite{monodepth2}   &   0.125  &   1.514  &   7.927  &   0.195  &   0.849  &   0.950  &   0.980\\
~& HR-Depth~\cite{lyu2020hr}   &   0.131  &   1.504  &   8.023  &   0.199  &   0.828  &   0.949  &   0.982\\
foggy & CADepth~\cite{yan2021channel}   &   0.126  &   1.375  &   7.585  &   0.187  &   0.845  &   0.956  &   0.986  \\
~ & DIFFNet~\cite{zhoudiffnet}  &   0.111  &   1.232  &   7.047  &   0.169  &   0.869  &   0.966  &   0.989  \\
~& \bf{MonoViT (ours)}           &   \bf{0.096}  &   \bf{0.934}  &   \bf{6.313}  &   \bf{0.150}  &   \bf{0.893}  &   \bf{0.974}  &   \bf{0.993} \\
\hline
\hline
~& Monodepth2~\cite{monodepth2}  &   0.155  &   1.900  &   6.976  &   0.209  &   0.813  &   0.943  &   0.979\\
~& HR-Depth~\cite{lyu2020hr}   &   0.149  &   1.656  &   6.658  &   0.204  &   0.815  &   0.945  &   0.981\\
cloudy  & CADepth~\cite{yan2021channel}    &   0.147  &   1.811  &   6.785  &   0.201  &   0.832  &   0.948  &   0.981  \\
~  & DIFFNet~\cite{zhoudiffnet}   &   0.140  &   1.571  &   6.298  &   0.192  &   0.837  &   0.950  &   0.983 \\
~& \bf{MonoViT (ours)}            &   \bf{0.125}  &   \bf{1.300}  &   \bf{5.970}  &   \bf{0.177}  &   \bf{0.861}  &   \bf{0.958}  &   \bf{0.986} \\
\hline
\hline
~& Monodepth2~\cite{monodepth2}  &   0.240  &   3.339  &  11.040  &   0.301  &   0.591  &   0.857  &   0.952 \\
~& HR-Depth~\cite{lyu2020hr}    &   0.222  &   2.962  &  10.494  &   0.281  &   0.631  &   0.868  &   0.959\\
rainy & CADepth~\cite{yan2021channel}    &   0.221  &   3.072  &  10.681  &   0.277  &   0.632  &   0.879  &   0.963  \\
~ & DIFFNet~\cite{zhoudiffnet}   &   0.191  &   2.411  &   9.626  &   0.244  &   0.679  &   0.914  &   0.978\\
~& \bf{MonoViT (ours)}            &   \bf{0.169}  &   \bf{1.925}  &   \bf{8.604}  &   \bf{0.219}  &   \bf{0.733}  &   \bf{0.934}  &   \bf{0.985} \\
\hline
\hline
~& Monodepth2~\cite{monodepth2}  &   0.155  &   1.740  &   6.744  &   0.214  &   0.819  &   0.941  &   0.977\\
~& HR-Depth~\cite{lyu2020hr}    &   0.153  &   1.546  &   6.505  &   0.212  &   0.812  &   0.942  &   0.978\\
sunny & CADepth~\cite{yan2021channel}  &   0.145  &   1.518  &   6.485  &   0.202  &   0.827  &   0.949  &   0.982  \\
~ & DIFFNet~\cite{zhoudiffnet}  &   0.142  &   1.457  &   6.165  &   0.197  &   0.835  &   0.950  &   0.982 \\
~& \bf{MonoViT (ours)}         &   \bf{0.130}  &   \bf{1.266}  &   \bf{6.109}  &   \bf{0.186}  &   \bf{0.851}  &   \bf{0.956}  &   \bf{0.985} \\
\hline

\end{tabular}}
\caption{\textbf{Results on DrivingStereo Dataset~\cite{yang2019drivingstereo}}. Models trained on KITTI with $640\times192$ images and tested on four different complex scenarios (\textit{foggy}, \textit{cloudy}, \textit{rainy} and \textit{sunny}).}
\label{tab:ds}
\vspace{-0.4cm}
\end{table}

\textbf{Results on DrivingStereo.} Additionally, to further evaluate the generalization capacity of MonoViT, we also test it under four different weather conditions, including \textit{foggy}, \textit{cloudy}, \textit{rainy} and \textit{sunny}, from the DrivingStereo~\cite{yang2019drivingstereo} dataset. In Tab.~\ref{tab:ds}, we collect the output of this evaluation for MonoViT and SoTA frameworks. Any model has been trained on KITTI and tested on DrivingStereo without any re-training or fine-tuning. Once again, MonoViT performance always results vastly superior to any CNN competitor. In this case, the margin is even higher than what was observed for KITTI and Make3D. This fact further suggests that the ViT encoder used within our framework dramatically affects the generalization capacity of the whole depth network, thanks to the long-range relationships among features modeled by the Transformer blocks themselves.

\begin{figure}[t]
\begin{center}

 \includegraphics[width=0.9\linewidth]{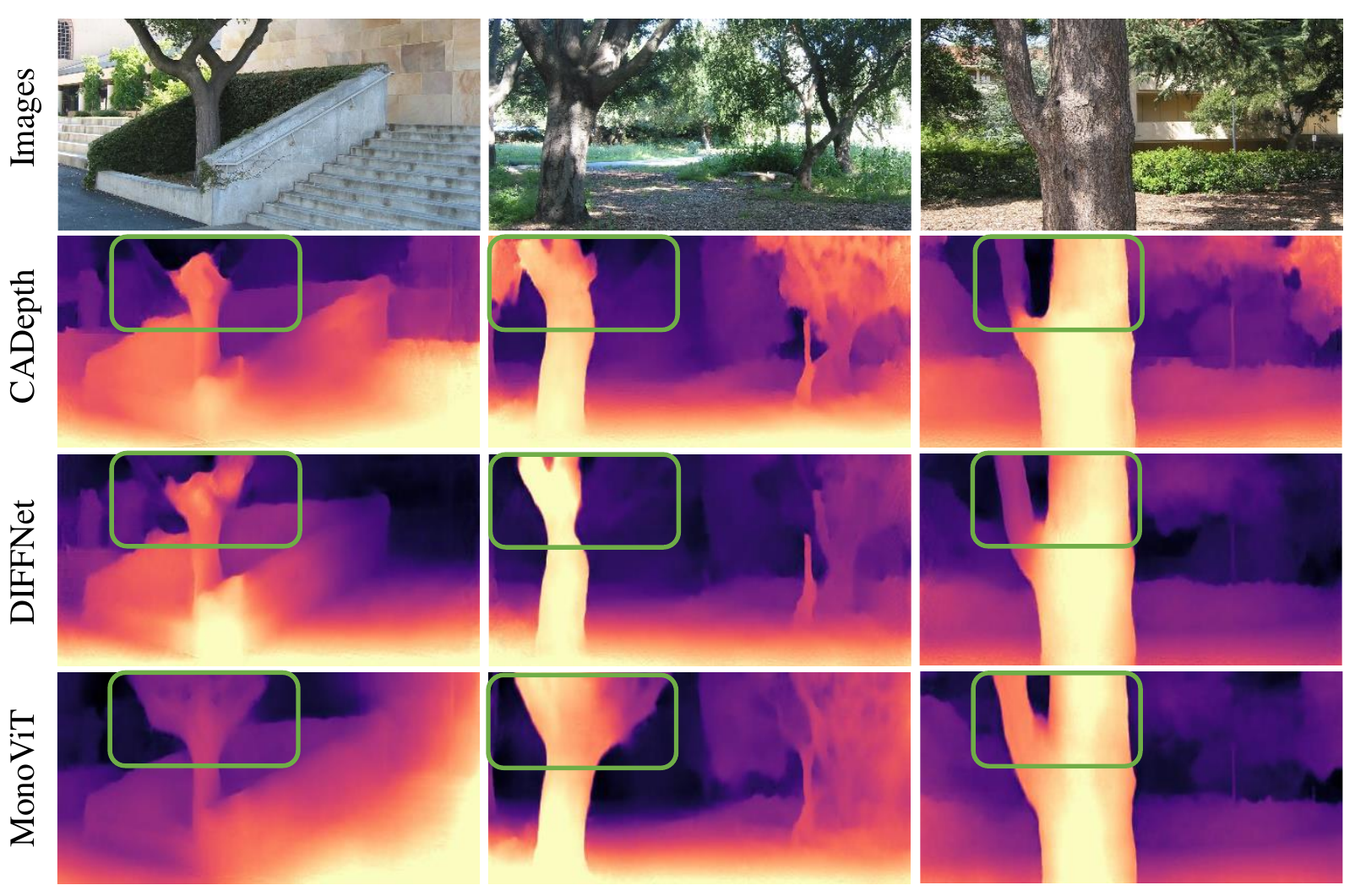}
\end{center}
\vspace{-0.4cm}
   \caption{\textbf{Qualitative comparison on the Make3D dataset~\cite{saxena2008make3d}.} Predictions by CADepth~\cite{yan2021channel}, DIFFNet~\cite{zhoudiffnet} and our MonoViT.}
\label{fig:m3d}
\vspace{-0.4cm}
\end{figure}

\begin{figure}[t]
\begin{center}
 \includegraphics[width=0.9\linewidth]{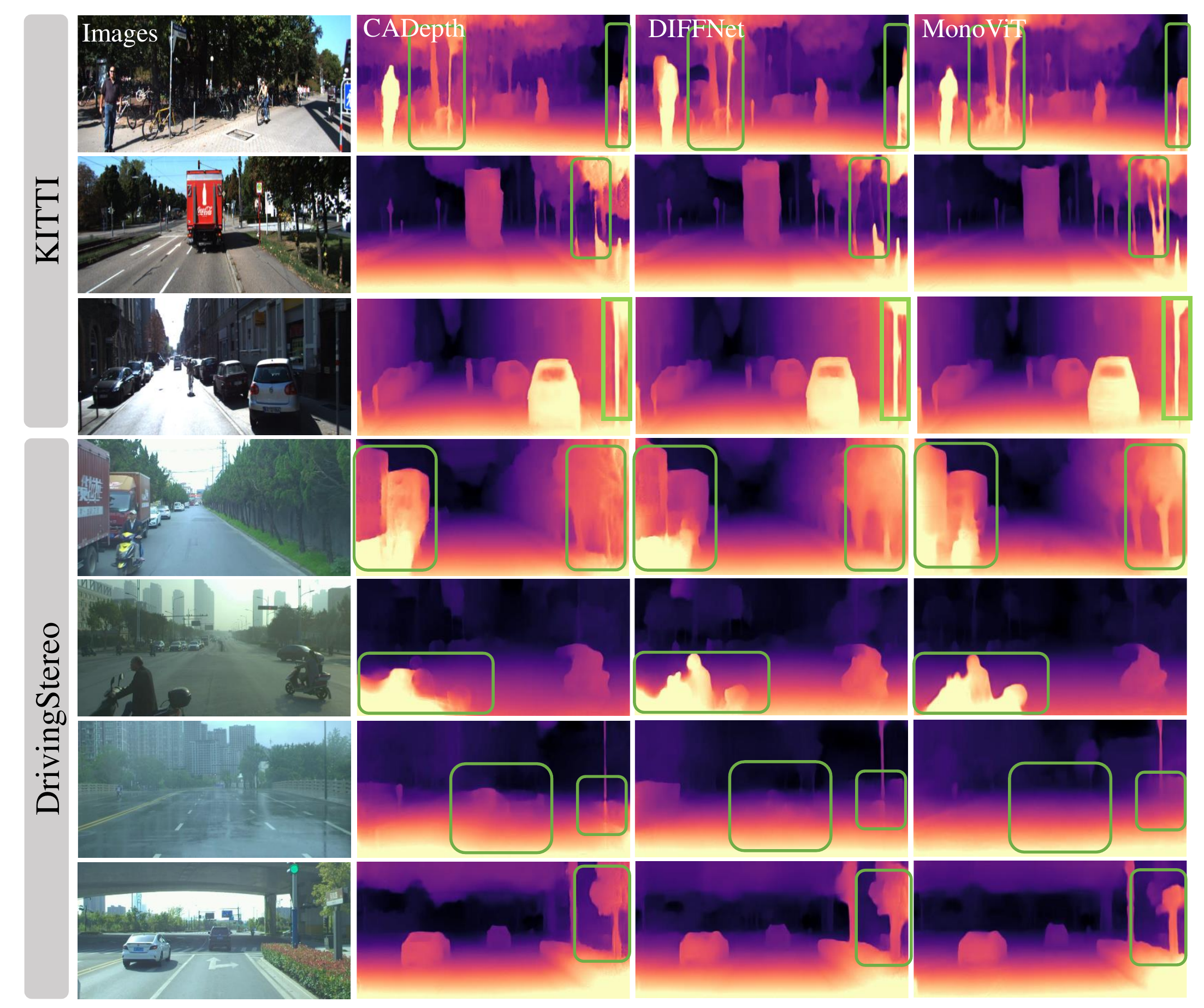}
\end{center}
    \vspace{-0.4cm}
   \caption{\textbf{Qualitative comparison on KITTI~\cite{kitti} (top) and DrivingStereo~\cite{yang2019drivingstereo} (bottom).} Predictions by CADepth~\cite{yan2021channel}, DIFFNet~\cite{zhoudiffnet} and our MonoViT.}
   \vspace{-0.2cm}
\label{fig:vkd}
\end{figure}

\textbf{Qualitative results.} Fig. \ref{fig:m3d} reports some qualitative examples from the Make3D dataset, with MonoViT being able to model the structures of objects more accurately than its competitors. Fig.~\ref{fig:vkd} shows a further qualitative comparison between MonoViT and SoTA frameworks on some challenging images from KITTI (top) and some even more challenging samples from DrivingStereo (bottom). For both datasets, we notice that MonoViT can effectively model the foreground and background because of the global receptive field, resulting in more precise, finer-grained estimation.

\subsection{Ablation study}
Finally, to further validate the effectiveness of our depth architecture, we report an ablation study in Tab. \ref{tab:ablation}. 
comparing the results yielded by MPViT variants (tiny, xsmall, small, base) and different recent Transformer encoders, SwinT-tiny \cite{liu2021swin} and PVT \cite{wang2021pyramid} (which counts a similar number of parameters compared to MPViT-small) on top, also by reporting the amount of parameters and FPS for each. Benefiting from the combination of CNNs and Transformers, the MPViT backbone outperforms the other two SoTA pure Transformer backbones (SwinT \cite{liu2021swin}, PVT \cite{wang2021pyramid}) and the pure CNN one (ResNet34~\cite{resnet}) in the self-supervised monocular depth estimation task.
Besides, we assess the impact of the different modules on bottom, like the Atten. Block in the decoder and the CNN path/Transformer path in the ``Joint CNN \& Transformer Layer" (Fig. \ref{fig:vit}). As shown in the table, both CNN path, Transformer path and Atten. Blocks play an important role in the architecture.

\begin{table}[t]
\centering
\scalebox{0.68}{
\setlength
\tabcolsep{3pt}{
\footnotesize
\begin{tabular}{l |c | c | c c c c| c c c}
\hline

\multirow{2}{*}{Backbone}& \multirow{2}{*}{Params$\downarrow$}& \multirow{2}{*}{FPS$\uparrow$}  &\multicolumn{4}{c|}{lower is better} & \multicolumn{3}{c}{higher is better}\\
 & & &Abs Rel$\downarrow$& Sq Rel$\downarrow$&  RMSE$\downarrow$ & RMSE log $\downarrow$& $\delta_{1} \uparrow$ & $\delta_{2} \uparrow$ & $\delta_{3} \uparrow$\\
\hline 
ResNet34~\cite{resnet} & 27M & \textbf{42}  & 0.108 & 0.780 & 4.622 & 0.183 & 0.884 & 0.963 & 0.983\\
SwinT-tiny \cite{liu2021swin} & 34M & 41  & 0.101 & \bf{0.698} & 4.404 & 0.177 & 0.894 & 0.966 & \bf{0.984}\\
PVT-small\cite{wang2021pyramid} & 30M & 38   & 0.106 & 0.801 & 4.648 & 0.184 & 0.887 & 0.961 & 0.982\\
MPViT-tiny & \textbf{10M} & 24  & 0.102 & 0.733 & 4.459 & 0.177 & 0.895 & 0.965 & \bf{0.984}\\
MPViT-xsmall &13M & 24  & 0.101 & 0.738 & 4.402 & \bf{0.175} & 0.898 & \bf{0.967} & \bf{0.984}\\
MPViT-small &27M & 18   & \bf{0.099} & 0.708 & \bf{4.372} & \bf{0.175} & 0.900 & \bf{0.967} & \bf{0.984}\\
MPViT-base & 78M & 15   & 0.100 & 0.747 & 4.427 & 0.176 & \bf{0.901} & 0.966 & \bf{0.984}\\
\hline
 MPViT-small &27M & 18   & \bf{0.099} & \bf{0.708} & \bf{4.372} & \bf{0.175} & \bf{0.900} & \bf{0.967} & \bf{0.984}\\
w/o CNN Path &- & -  & 0.114 & 0.929 & 4.821 & 0.190 & 0.879 & 0.959 & 0.981\\
2 Trans. Path &- & -  & 0.107 & 0.801 & 4.590 & 0.182 & 0.889 & 0.963 & 0.983\\
1 Trans. Path &- & -  & 0.120 & 0.876 & 4.799 & 0.196 & 0.864 & 0.956 & 0.981\\
w/o Trans. Path &- & -  & 0.127 & 0.931 & 4.972 & 0.203 & 0.850 & 0.951 & 0.980\\
w/o Atten. Block &-  & -  & 0.101 & 0.772 & 4.465 & 0.177 & 0.898 & 0.965 & 0.983\\
\hline
\end{tabular}}
}
\caption{\textbf{Ablation study on KITTI.} Trans. refers to Transformer. Input is 640$\times$192, runtime measured on RTX 3090 GPU. Encoders are pre-trained on ImageNet. 
}
\label{tab:ablation}
\vspace{-0.4cm}
\end{table}

%% file: 6conclusion.tex
\section{Conclusion}

This paper proposed MonoViT, a new architecture for self-supervised monocular depth estimation.
By combining both convolutions and Transformers block inside the network encoder, MonoViT can model the local and global context of images jointly, overcoming existing solutions based on CNNs.
Our proposal vastly and consistently outperforms the SoTA on the KITTI dataset. Moreover, experiments on Make3D and DrivingStereo datasets show that MonoViT achieves better generalization performance than SoTA architectures for self-supervised depth estimation. 

\small \textbf{Acknowledgements:} This work was supported in part by national Key  Research \& Development Program - National Key Research and Development Program of China (2021YFB1714300), National Natural Science Fund for Distinguished Young Scholars (61725301), Programme of Introducing Talents of Discipline to Universities (the 111 Project) under Grant B17017, Innovation Research Funding of China National Petroleum Corporation (2021D002-0902) and Shanghai AI Lab.